# Adding LLMs to the psycholinguistic norming toolbox: A practical guide to getting the most out of human ratings

Javier Conde [1], María Grandury [1], Tairan Fu [2], Carlos Arriaga [1], Gonzalo Martínez [1], Thomas Clark [3], Sean Trott [4], Clarence Gerald Green [5], Pedro Reviriego [1], Marc Brysbaert [6]

[1] Information Processing and Telecommunications Center (IPTC), Universidad Politécnica de Madrid (Spain)
[2] Politecnico di Milano (Italy)
[3] Massachusetts Institute of Technology (United States)
[4] Department of Cognitive Science, University of California (United States)
[5] Faculty of Education, University of Hong Kong, (Hong Kong)
[6] Department of Experimental Psychology, Ghent University (Belgium)

Correspondence:   Javier Conde
Information Processing and Telecommunications Center
Universidad Politécnica de Madrid
28040 Madrid, Spain
javier.conde.diaz@upm.es

# Abstract

Word-level psycholinguistic norms lend empirical support to theories of language processing. However, obtaining such human-based measures is not always feasible or straightforward. One promising approach is to augment human norming datasets by using Large Language Models (LLMs) to predict these characteristics directly, a practice that is rapidly gaining popularity in psycholinguistics and cognitive science. However, the novelty of this approach (and the relative inscrutability of LLMs) necessitates the adoption of rigorous methodologies that guide researchers through this process, present the range of possible approaches, and clarify limitations that are not immediately apparent, but may, in some cases, render the use of LLMs impractical.

In this work, we present a comprehensive methodology for estimating word characteristics with LLMs, enriched with practical advice and lessons learned from our own experience. Our approach covers both the direct use of base LLMs and the fine-tuning of models, an alternative that can yield substantial performance gains in certain scenarios. A major emphasis in the guide is the validation of LLM-generated data with human "gold standard" norms. We also present a software framework that implements our methodology and supports both commercial and open-weight models.

We illustrate the proposed approach with a case study on estimating word familiarity in English. Using base models, we achieved a Spearman correlation of 0.8 with human ratings, which increased to 0.9 when employing fine-tuned models. This methodology, framework, and set of best practices aim to serve as a reference for future research on leveraging LLMs for psycholinguistic and lexical studies.

# 1. Introduction

The rapid growth of Large Language Models (LLMs) has provided scientists with a range of new tools. In psycholinguistics specifically, one of the most important applications of LLMs has been the estimation of surprisal, the degree to which a word is expected based on the preceding context. LLM-based estimates of surprisal have provided more informative estimates of word predictability (Cevoli et al., 2022; Wilcox et al., 2023), outperforming human sentence completions (known as cloze probabilities; Staub, 2025).

This article discusses the generative capacity of LLMs: the creation of new, artificial data to simulate human performance without having to collect data directly from people (Long et al., 2024). This generative capacity has been used, for instance, for the generation of synthetic data that can be used to train new versions of LLMs. Evidence suggests that the generative capabilities can also be used for other purposes, such as augmenting human datasets for scientific investigation. LLMs provide useful (albeit imperfect) estimates for various lexical properties with simple prompts, which would otherwise require expensive data collection from humans.

A limitation of LLMs for psycholinguistics is their lack of explainability (Jain and Wallace, 2019). This is primarily due to the architecture of the models, which are based on complex neural networks, making it very difficult to understand how the model arrives at its output. In addition, the best-performing models currently available are commercial models that are not open to scientific scrutiny (Bommasani et al., 2024). As a result, LLMs are a kind of "black box" that provides intriguing and potentially useful information, without users knowing how the output is generated.

It should be noted that the inscrutability of LLMs is not specific to artificial intelligence. Psycholinguists have been collecting data on human performance for over 150 years without knowing exactly how the human brain generates the data. When researchers ask participants to rate words on familiarity, they are not solely (or even primarily) interested in how the ratings are made, but whether the ratings are a valid measure of word knowledge and can be used, for example, to predict the difficulty of a text. For many purposes (such as norming experimental stimuli), predictive power is as important as being able to explain the underlying processes (Yarkoni & Westfall, 2017). At the same time, if a researcher's interest is obtaining human ratings, human responses are at least (definitionally) *face valid* measures; the same is not necessarily true of LLM-generated responses, and the inscrutability of LLMs is one factor that makes assessing the validity of synthetic data challenging. Still, LLMs may be useful tools for obtaining estimates of word characteristics, building on past approaches (Thopmson & Lupyan, 2018).

As in any field of research, collecting psycholinguistic data from LLMs requires knowledge of how best to query the models and what steps to take to obtain the most useful output. We have found that over the past two years, we have built up a considerable amount of implicit knowledge about how to access an LLM (e.g., using an application programming interface instead of a web interface), which prompts to use (prompt engineering), how to optimize estimates with fine-tuning, and how to avoid common pitfalls. In this article, we make this implicit knowledge explicit as a practical guide to using LLMs for psycholinguistic norming. We hope that this guide will be useful to researchers in the field, enhance reproducibility, and spark further discussion and iterative improvements regarding best practices. Further, because the focus of this guide is on *practical usage*, we do not discuss theoretical underpinnings of text-based LLMs (e.g., the distributional semantic hypothesis) in detail, except when relevant for discussing concerns with the validity of LLM-generated data.

This manuscript is divided in two main parts. In the first one (section 2-8) we briefly discuss the nature of LLMs, explain how generating estimates of word features using LLMs, based on best practices, as this has consequences on what type of information can be obtained and how best to obtain it. In the second part (Appendix A) we present our framework[1] and a step-by-step tutorial for obtaining norms from LLMs with and without fine-tuning.

## 2. Large Language Models

### 2.1. Introduction to LLMs

The widespread use of generative AI began with the launch of ChatGPT 3.5 in November 2022, based on OpenAI's GPT 3.5 model. GPT stands for Generative Pre-trained Transformer. It is a type of language model developed using a neural network architecture known as a transformer, which enables efficient text processing by capturing relationships between a large set of words. The term "generative" indicates that the model can be used to produce new text, rather than simply processing or classifying existing text. 'Pre-trained' means that it has been trained on large amounts of text (in the case of the largest models, on billions or trillions of words). This pre-training allows the model to learn which words (and types of words) are most likely to occur in different contexts, often reflecting underlying syntactic rules or semantic associations (Trott & Bergen, 2023; Jones & Bergen, 2024). Once trained, the model is able to predict tokens (fragments of words).

A word is made up of one or more tokens, with fewer tokens assigned to more frequent words (Arnett & Chang, 2024). Models work with tokens instead of words because the size of the neural network would be much larger if it had to deal with all possible words and their variations, especially given that most models are multilingual. Training a *tokenizer* (i.e., building the dictionary of tokens) is a preliminary phase before training the models.

---

[1] https://github.com/WordsGPT/psycholinguistics_framework

Token dictionary sizes typically range in the tens or hundreds of thousands and are capable of generating any word in the languages supported by the LLM. As an example, the token dictionary used in Llama3 consists of 128 thousand tokens (Grattafiori et al., 2024). The word "picture" is composed by a single token, but the word "picnic" by two tokens "pic" and "nic".

Most tokenizers have been designed to optimize English, which makes generating the same text in a different language more expensive, slower, and more energy-consuming, as it requires a larger number of tokens (Petrov et al., 2023), although over successive generations, this limitation has been reduced[2]. Moreover, there is some evidence that tokenization strategies are biased towards certain kinds of morphological and orthographic systems (Arnett & Bergen, 2024; Arnett & Rivière, 2024).

An English bias is also present for pre-training, where the prevalence of English datasets leads to higher quality in that language (Huang et al., 2025). Some studies even show that models may "translate" input tokens to English in intermediate layers for processing (Wendler et al., 2024). Techniques such as automatic translation of datasets have been used to increase the training and evaluation of LLMs (Grattafiori et al., 2024; Achiam et al., 2024) to reduce this gap. However, this process shows limitations inherent in the translations themselves (Plaza et al., 2024).

Once a language model has been trained, the output is a set of parameters, mostly weights, (hundreds of billions in the largest models) that store the model's knowledge. These weights are static when the model is used in *inference* mode, meaning that the model does not continue to learn after the training process is complete, but rather applies what has been learned to generate responses or perform tasks.

Although models do not learn while in use, their responses can be guided by information in the *context*. The *context window* refers to the number of tokens that a model can keep track of in a conversation. The context window has increased significantly over the years, from a few thousand tokens to up to one million tokens in the Gemini models (Comanici et al., 2025). However, having an extensive context window does not guarantee that the model will track all information in the context equally accurately, as attention tends to degrade with distance in the context, causing certain parts to receive more attention than others (Comanici et al., 2025).

## 2.2. Explainability and Alignment of LLMs

Although the LLM produces output by predicting tokens, the output is not just a single token, but rather a probability distribution over all tokens in the dictionary. The final selection of the output token is made through a weighted sampling based on the probability assigned to each token. One option is to always select the token with the highest probability, making the model deterministic since it (almost[3]) always produces the same response to the same input. However, the typical configuration of the models introduces randomness, which allows for more varied and natural responses, something fundamental to simulating the flexibility of human language. (Note, however, that while introducing randomness does increase the *novelty* of generated text, this does not necessarily reflect increased "creativity" (Peppercorn et al., 2024)).

---
[2] https://openai.com/index/hello-gpt-4o/

The probability distribution of the output tokens is expressed as log-probabilities, or logprobs. In principle, with sufficient computational resources, one could produce a high-probability string of length *n* by exploring every possible combination of token sequences in the model's vocabulary. However, the high number of combinations at each step makes this task computationally infeasible, turning explainability into one of the greatest current challenges of LLM.

Explainability consists of understanding how all the model's parameters interact to generate coherent linguistic behavior. One central challenge to explaining or interpreting LLMs is the sheer size of the models: with billions or even hundreds of billions of parameters—and infinite numbers of possible input sequences—identifying the mapping between each parameter and each kind of input is computationally intractable. Researchers have attempted to gain purchase on this question by designing targeted experiments that allow them to probe the inner workings of the models in response to some experimental manipulation (Bricken et al. 2023, Templeton et al., 2024), though interpretability remains an open challenge.

It may be tempting for users to ask an LLM how it arrived at its output. Unfortunately, this is not necessarily an effective strategy (Hu & Levy, 2023). LLMs may not be capable of introspecting on the mechanisms that give rise to their own behavior (Song et al., 2025); in general, the fidelity of a model's explanation for its behavior remains an open question (Turpin et al., 2023). In many cases, an LLM might give a textbook example of what it should take into account, regardless of what is happening within the model.[4]

Since explainability is currently an unsolved problem, one of the key goals in LLM development is to ensure alignment with human interaction patterns. This alignment not only involves producing grammatical and coherent responses, but also responses that are perceived as useful, safe and socially acceptable. Ensuring alignment is important to prevent LLMs from negatively interfering with the cognitive abilities of people (Niu et al., 2024).

---

[3] Even if the temperature is set to 0 to select always the token with the highest probability and with the same configuration of parameters, small differences may still occur in the output of LLMs: https://152334h.github.io/blog/non-determinism-in-gpt-4/

[4] To illustrate, we invite readers to ask an LLM to estimate familiarity for a set of words and then ask to the model how it arrived at the estimates.

The lack of explainability, combined with the presence of dependencies and randomness in LLMs, means that output can vary greatly from session to session, depending on how the AI model is configured and how it is handled. This requires a systematic approach to eliciting responses if LLMs are to be adopted into the psycholinguistic toolkit.

## 3. Obtaining outputs from Large Language Models

There are (roughly) two approaches to eliciting responses from an LLM: through web interfaces or via APIs (Application Programming Interfaces). The choice of approach has implications for control, functionality, and customization.

Web interfaces include chatbots that are accessible via browsers. They are designed for direct communication with users and enable natural conversations. Modern chatbots are not limited to text, but also support audio, images, or video and can generate output in any of these formats. The main advantage is that they are extremely easy and intuitive to use. For example, you can ask a browser to estimate the familiarity of a series of words.[5]

A limitation of most popular web interfaces is that users often cannot adjust the parameters used to elicit text from the model. This has two consequences. First, the output is likely to vary slightly from session to session (see Temperature below), and second, the estimates will be influenced by the questions asked and answers given before (due to the window in which the LLM operates). The estimates are therefore likely to differ if the same word is presented in different word sequences. Another limitation is that ease of use becomes a burden when several thousand or millions of prompts have to be entered manually. Finally, many models hosted on a chat interface have been given a "system prompt" that also influences their behavior, which may not be publicly available (see discussion of closed vs. open models in Section 4).

Interaction with LLMs is also possible via APIs, which allow you to communicate with the models using programming languages. This requires knowledge of (the basics of) a programming language (such as Python or R) and the purchase of an API key. API access is generally managed via a pay-per-use model, with costs depending on the number of input and output tokens. With an API, you can retrieve the information from your computer code. This not only automates the search for information, but also allows you to change the configuration of the model parameters, which directly affects the behavior of the model. These settings allow you to influence the style, coherence, or creativity of the generated responses. Among the most commonly used configurable parameters are temperature and top-k.

---

[5] Again, the reader is invited to do so. For example: "Can you give me the familiarity (from 1 = very unfamiliar to 7 = very familiar) of the following words: psycholinguistic, child, aardvark, ambitious, and zucchini."

Temperature controls the degree of randomness in the responses by changing how tokens are sampled from a model's probability distribution over possible completions in a given context. Low temperature values make the sampling strategy more deterministic and predictable, while higher values yield more varied and unpredictable responses. Ideally, a temperature of 0 should make the system completely predictable, but this is not the case for many commercially available models, possibly because of how sampling is divided across GPUs. However, the output of two queries with temperature 0 will usually be highly correlated.

The top-k parameter limits the number of options (output tokens) that the model considers when performing the weighted sampling. When top-k = 1, the model always chooses the single, most likely answer. To produce norms, top-k is typically set to 1 or the result is the weighted mean of the possible answers based on the logprobs.

There are other parameters, such as top-p, frequency penalty, or presence penalty, that can be used to control the model's choice of the next token, but we have not explored the utility of modifying these parameters from their default values.

A final, very important parameter for generating information is the extent to which the model is influenced by previous inputs. If stable estimates are required, you must ensure that each elicitation is a new, independent query. This limits the possibilities for working with batch tests (i.e., giving the model a list of words and asking it to generate an estimate for each of the words). A good way to test whether the estimates are influenced by previous estimates is to use two different permutations of the list for which you need estimates. If the two estimates correlate less than 0.96, it is likely that there are dependencies in the estimates you obtain (or that the range of estimates is too small to obtain reliable differences in estimates).

## 4. Which LLM to use?

There is a large and rapidly growing variety of LLMs to choose from, many of which can be found at companies' websites (for commercial models) or public repositories such as Hugging Face[6] (for open-weight models).

---

[6] https://huggingface.co/

The first selection criterion is whether the LLM is commercial, open-weight, or research-based. **Commercial** models do not share the model weights (or the training data), so all interactions must take place via the model owner's servers. Examples of this type are the GPT or Gemini model series from OpenAI and Google. **Open-weight** models, on the other hand, do publish the weights, allowing people to download the model and run it in inference mode using public or private compute infrastructure (which is not that of the model owner). Although the weights are accessible, the training data and full details of the development process remain hidden. Models such as LLaMA (Meta) or DeepSeek (DeepSeek AI) are examples of this category. **Fully-open** models give users complete control over the training material and training regime. Platforms and models like Pythia (Biderman et al., 2023), OLMo 2[7] or Apertus[8] provide a suite of LLMs with all the data and process public to be used in interpretability experiments. Other advantages include that they are free to use and that one can study (and manipulate) the model's operation. The main disadvantage of these models at present is that they do not achieve the performance level of commercial and open-weight LLMs. At the same time, a chief concern with proprietary models (as well as open-weight models, for which the training data is unknown) is *data contamination*: the possibility that they have been trained directly on data used to evaluate the model. This makes evaluating the capabilities of those models extremely difficult (if not impossible). Another major limitation of commercial models specifically is reproducibility: because much of the model workings are hidden behind an API, their behavior may vary across repeated experiments, making it impossible to *exactly reproduce* their behavior. See additional work (Liesenfeld et al., 2024; Hussain et al., 2024) for a much more detailed consideration of the concerns relating to commercial, open-weight, and fully-open models.

The second selection criterion is the quality of the model. If the primary aim is to obtain and publish psycholinguistic norms that approximate human judgments, the performance of the model may outweigh concerns about openness and justify the use of a closed-source, proprietary model. However, if differences between models are small and have little practical value, other considerations come into play, such as the increasing commitment in psycholinguistics to open science and inclusive practices. At present, and from our experience in these tasks, proprietary models offer significantly better performance, but as research progresses, we hope that open-source alternatives can be prioritized. Cost-benefit considerations matter for the use of public research funds, as does the relationship between public funds and private enterprises. If an open-source model yields correlations within the acceptable range with human norms, we recommend the use of that model for the reasons stated above (reducing the possibility of data contamination and allowing for direct, exact reproducibility).

---

[7] https://allenai.org/olmo
[8] https://ethz.ch/en/news-and-events/eth-news/news/2025/09/press-release-apertus-a-fully-open-transparent-multilingual-language-model.html

A typical approach to evaluating different models is comparing their output to validation datasets (Conde et al., 2025a). A validation dataset consists of human data for the variable of interest. For example, if one wants to obtain LLM estimates of the age at which a word was learned (age of acquisition or AoA), one looks for a set of human AoA ratings with which to compare the output of the LLM. Ideally, multiple human sets are available so that one can not only see how well the LLM output correlates with the human ratings, but also to what extent this correlation is comparable to the correlations between different sets of human ratings, i.e., *inter-annotator agreement*. An inter-annotator correlation score of about 0.8 or 0.9 is generally considered reliable. Moreover, the degree of inter-human correlation represents a threshold for humanlike model performance: if a model produces values for a particular psycholinguistic norm whose correlation with human data is indistinguishable from the inter-human correlation, it suggests that individual model judgments may be substitutable for individual human judgments, in the sense that replacing a random human judgment with a model judgment would not meaningfully change the average judgment for that item. In some cases, model outputs have shown even higher correlations with the human average than have individual human judgments, suggesting that model judgments may capture a partial "wisdom of the crowd" (Trott, 2024b).

If no human ratings are available for a domain of interest, we strongly recommend collecting them for a representative sample of several thousand words. This allows researchers to perform the necessary validation checks; an additional advantage is that the human ratings have not yet influenced the LLM estimates, i.e., there is no possibility of direct contamination given that the ratings are entirely new. As noted previously, one concern with LLMs (especially those whose training data is not publicly available) is that their output can be influenced by datasets available on the internet. In these cases, the LLM might produce good estimates for words in the dataset on the internet, but not for new words. When validating the output of an LLM on an established dataset, one must indeed always take into account that good performance on that dataset cannot necessarily be generalized to other words. Therefore, it is always a good idea to perform some additional checks on the quality of the estimates obtained with words that are not yet available in the literature. **In general, we cannot recommend publishing LLM-generated norms if the methodological procedure for generating those norms has not been validated with human data**.

Until 2025, we found that GPT-4o outperformed other models tested on the datasets we explored (constituting a range of languages and linguistic constructs), but this may not remain the case: other models are likely to improve, new models will be introduced, and the company developing GPT may make changes that make the model less suitable for estimating psycholinguistic word characteristics. Further, any given model can be 'improved' by fine-tuning that model on a pool of existing psycholinguistic norms (see below).

The best way to ensure that you are working with a good model and that you have done everything correctly is to compare the model's output with (new) human data. This should be a basic check when using LLM estimates. Beyond correlating judgments with a new gold standard, researchers can conduct *substitution analyses* to ask whether LLM-generated norms provide equivalent predictive validity, e.g., of reaction time or other dependent variables of interest (Trott, 2024 a). Such analyses might also inform questions about the construct validity of LLM-generated norms: if synthetic norms correlate strongly with human norms, *and* they predict other variables of interest in similar ways as do human norms, it suggests they might serve a useful function for psycholinguists. Researchers can also investigate LLM-generated norms for the presence of *systematic biases*. For example, Trott (2024a) found that GPT-4 systematically overrated the similarity of antonyms, and also exhibited particularly poor performance when rating the similarity of *concrete* (as opposed to *abstract*) word pairs. In some cases, the presence of systematic errors (i.e., relative to human judgments) may suggest that the mechanisms underlying LLM-generated judgments are different in important ways from those underlying human judgments. Thus, while the direct correlation with human ratings is probably the most important indicator of suitability, other analyses may help researchers triangulate the question of whether LLM-generated judgments are appropriate to use in a given context.

## 5. Prompt engineering

The next decision that will affect the quality of your LLM estimates is the prompt you use to elicit a given output. Some prompts are more effective at eliciting human-aligned judgments than others. Finding the best prompt is called *prompt engineering*. Prompt engineering involves selecting the target language, context, instructions, and so on.

Prompt engineering is an active research area, but there is evidence that small variations in prompts can produce different responses. For example, specifying the desired tone ("write in a formal style"), indicating the output format ("return a numbered list"), or setting the model's role ("act as a language expert") could affect the quality of the response. One widely used and seemingly reliable technique is called few-shot prompting, in which a model is provided with one or more examples of the task; typically, these examples help the model produce answers in the appropriate format that are more aligned with the user's goals.

When thinking of a good prompt, it is good to keep in mind that LLMs are trained to produce the most likely next token. Ideally, a good prompt should point the model directly to that token. For instance, we found that the following prompt worked well to get concreteness estimates for words (Martínez et al., 2024):

> Could you please rate the concreteness of the following word on a scale from 1 to 5, where 1 means very abstract and 5 means very concrete? Examples of words that would get a rating of 1 are essentialness, although and hope. Examples of words that would get a rating of 5 are bat, frangipane, and blackbird. The word is: [insert word here]. Only answer a number from 1 to 5. Please limit your answer to numbers.

This prompt directly asks for concreteness estimates between 1 and 5; it explains what 1 stands for and what 5 stands for. However, it does assume that the model 'understands' concreteness, as no definition is given. When it is unclear whether a model will understand the construct in question, we recommend giving an explicit definition of the construct in the prompt, e.g., by starting the prompt with the sentence "Concreteness is a measure of how concrete or abstract something is. A word is concrete if it represents something that exists in a definite physical form in the real world. In contrast, a word is abstract if it represents more of a concept or idea." (Scott et al., 2019).

The prompt we provided also includes three examples of words that would get an estimate of 1 or an estimate of 5, which help scaffold its responses. When these are not given, the model is left on its own to decide what the extremes are. Prompting without examples is called zero-shot prompting.

The advantage of zero-shot prompting is that you rely entirely on the LLM's pre-existing knowledge and its understanding of natural language instructions. You do not influence the output in any way. The disadvantage is that the model may work with suboptimal choices, for instance using too narrow a range, so that the estimates are squeezed between 2 and 4. The only way to see whether giving examples is helpful is to compare the output of the different prompts to a set of human ratings (or other information you have about the variable).

The prompt we used also instructs the model to provide only numbers. This is necessary because otherwise the model may produce irrelevant output (e.g., comments on the given word or on the task in general). Even with explicit instructions, it is necessary to always check the output for incidental hallucinations, i.e., responses that are not to be expected based on pragmatic human conversation.

As explained in the previous section, we repeat the prompt for each word in an entirely new API call to avoid the risk of dependencies in the answers. Nevertheless, it is good practice to occasionally check that the estimates do not differ depending on the order in which the words are given within a session, and that an estimate is given for each word.

Occasionally, we have observed that the LLM does not provide an estimate for a word (e.g., a taboo word or even a random word from the list) and immediately moves on to the next word. This carries the risk that the words and estimates will not match if the pipeline is fully automated without cross-checks. One way to protect yourself against misalignment is to ask the model to provide both the word and the estimate by giving the command "The output format must be a JSON object. For example:  {"Word": "{word}", " Concreteness": // Concreteness of the word expressed as a number from 1 to 5}". This makes it easy to check whether estimates have been provided for all words and, if not, which words are missing. However, this can have disadvantages. In general (but not always), we have found that asking for additional information carries the risk of reducing the quality of the critical estimate, as there is no longer a direct link between the prompt and the output. When asking for additional information, it is a good idea to check that this does not affect the answers given.

The prompt we used was in English. An open question is what the best prompt is for other languages. A natural intuition is that prompts in the target language (Spanish, German, Chinese, etc.) would be better, as this increases the connection between the input and the desired output. However, we have occasionally observed that the output is better when the prompt is in English, even to obtain estimates for another language. One way to understand this is that the LLM has to rely on its prior training to interpret the prompt correctly (what do the words concreteness, essentialness, bat mean?). If a model is not well trained in the target language, it is possible that an English prompt cues more activations that are relevant to the construct of interest than a prompt in the target language. The only way to find out if this is the case is to again compare the output of the different prompts with a validation set.

A final point is the granularity of the output. In our prompt, we requested a single number between 1 and 5. One could argue that this output is too coarse to be useful and that a finer distinction is desirable. One way to achieve this is to increase the range of numbers (e.g., between 1 and 100) or to request output with decimals (e.g., two decimals). This may improve the fidelity of estimates, but increased fidelity is not guaranteed: we have noticed that the model sometimes only produces a subset of possible answers. For example, in a recent study generating AoA for English words, GPT-4o tended to prefer whole number estimates, even when instructed that it could include decimal places and the model was fine-tuned to 2000 ratings of Kuperman et al. (2012) that included decimal places. One factor that could influence a model's behavior here is the number of tokens associated with a given rating: generally, a number with decimals (4.25) will consist of more tokens than a whole number (4).

A better alternative, when it is available, is directly inspecting the probability associated with different answers. If the logarithmic probabilities (logprobs) associated with different tokens are available, researchers can calculate a weighted average using those logprobs. For example, a model's most likely estimate for the familiarity of zucchini may be 4 (out of 5), but the logprobs may indicate that the probability of answer 3 is 0.1, that of answer 4 is 0.5, and that of answer 5 is 0.4. The weighted average is then 0.12*3 + 0.45*4 + 0.43*5

= 4.31. In this way, we have calculated estimates of word familiarity, concreteness, valence, and arousal (Conde et al., 2025a; Brysbaert et al., 2025). However, this does not always work when fine-tuning is applied, as we will see below. With an open-weight model provided through platforms like HuggingFace, the probability associated with any arbitrary token completion can be obtained, allowing researchers to obtain a probability distribution over the entire space of possible completions (if desired); this is yet another advantage of (relatively) open models.

## 6. Fine-tuning

LLMs are trained to predict the next token based on an input sequence. However, their popularity has grown with the emergence of chatbots. Chatbots do more than conveying information from the base model directly. They have been retrained within a second phase known as instruction-tuning. In this phase, the weights of the LLM are adjusted using specific datasets containing examples of questions and answers, allowing the model to adapt its behavior toward conversational interaction.

Fine-tuning is particularly effective in the final stages of model development, when all the knowledge from pretraining has already been acquired. At this point, the goal is not to learn new facts, but to adapt its output to specific tasks, modify its response style, adjust its tone, avoid harmful responses, and more. This personalization is possible because the process directly modifies some of the model's weights, so that the model becomes tailored to a specific task. Although fine-tuning alters some of the model's parameters, its ability to incorporate new knowledge is limited. Knowledge acquisition occurs primarily during the early stages of pretraining, when the model is exposed to vast amounts of text. Fine-tuning, by contrast, does not aim to expand general knowledge but to fine-tune the model's output for specific contexts or tasks.

The best-known example of fine-tuning is that of conversational models, but it is not the only one. Today, there are many other uses of the technique. For example, reasoning models are developed through an additional fine-tuning phase in which reinforcement learning with human feedback (RLHF) is applied, training the model to explore effective reasoning sequences. Fine-tuning can also be used to adapt a base model to specific tasks such as sentiment classification, summarization, or entity extraction. In these cases, the model becomes more competent at specific tasks thanks to the refinement of its behavior based on representative examples.

Our recent research suggests that fine-tuning can also be used to provide psycholinguistic estimates that are more in line with human judgment. Here, a model can be fine-tuned on a limited training set of words with human judgments, allowing researchers to ask whether this fine-tuning improves the quality of generated estimates for other words not in that fine-tuning set. For instance, Sendín et al. (2025) found that AoA estimates for Spanish words did not correlate as strongly with human judgments as the human judgments did with each other. The authors refined the model based on 2,000 human ratings and found that the correlation with human ratings increased to the same level as the correlations between the human ratings, even for words not included in the fine-tuning set. This suggests that fine-tuning may be particularly useful for languages that are underrepresented in the model's training or for response types that may be less "natural" to produce (like response time).

While fine-tuning should always be considered, some data indicates that it may be less helpful for some languages (like English) or tasks. For example, a recent study generating English AoA's investigated a trained model fine-tuned on 2000 ratings from Kuperman et al. (2012) and although fine-tuning increased correlations with human ratings, there was not much improvement over zero-shot untrained estimates for the same words (Green et al., 2025). This likely reflects the fact that the LLMs are overwhelmingly trained on English language data, or that this dataset is in the training dataset.

Fine-tuning works by first asking the model for an estimate and then providing feedback on the expected number. The feedback is used by the model to adjust the weights so that the output is closer to the expected output. The weights are readjusted sequentially for each word in the fine-tuning set. After fine-tuning, the adjusted model can be saved for future use.

Again, fine-tuning is a non-deterministic technique involving many variables (size of the training dataset, sampling strategy, model configuration, parameters, etc.). It is fair to say that we do not yet have a complete picture of what the best fine-tuning strategy is. How detailed can the feedback be (for example, can we use decimals for ratings from 1 to 5)? What are the most informative stimuli for fine-tuning a model? To what extent can the items we use skew the estimates of the other items? How large should the sample be? What information should the model provide (e.g., word + estimate or estimate only?) and what feedback is given to the model? These are all parameters that may or may not affect the quality of the fine-tuning and do not necessarily have to be the same for all languages and all variables that can be tested.

Fine-tuning LLMs is a promising tool for generating word features, as it can enable researchers to obtain useful information for an entire language based on a small subset of the data. In our recent studies (Sendín et al., 2025; Martínez et al., 2025), we have indeed found that it is possible to obtain useful response times for English lexical decisions after fine-tuning GPT-4o mini on RTs for 3000 words collected in a megastudy. Instead of collecting lexical decision times for tens of thousands of words, it is therefore possible to estimate them based on a sample of 3,000 words. This may be particularly interesting for populations that are difficult to reach and therefore cannot be tested in large-scale megastudies, or to try out different hypotheses to know which items are likely to yield the most information in tests with humans.

Another question is whether fine-tuning in a given language (e.g., English) improves performance in other languages (e.g., Spanish). This is especially important when the target language is relatively under-represented in a model's training data. In our research on familiarity estimates, we observed gradually deteriorating results for zero-shot queries from a model fine-tuned in English to Spanish, German, and Dutch. Future research could attempt to develop *theoretical principles* for when improvements from fine-tuning in one language should be expected to transfer to another language.

## 7. Pitfalls, Recommendations, and Open Questions

Throughout the guide thus far, we have discussed several common pitfalls researchers might encounter when using LLMs to generate psycholinguistic norms. In this section, we expand on these concerns, including our own recommendations for addressing them. We also discuss a number of open questions relating to the validation and use of LLM-generated norms.

### 7.1. Common pitfalls

First, as noted above, researchers may use approaches that do not allow sufficient experimental control, such as using the web interface for ChatGPT rather than the API; similarly, researchers may fail to *report* the precise parameters used to produce the outputs they analyzed, such as temperature and the exact prompt. Model openness relates to this concern: the use of a commercial model (such as GPT-4) often obscures the specific parameters involved in generating responses, which means that even if researchers report the *controllable* parameters used in their study, other researchers may fail to reproduce the exact responses they obtained. Commercial models also carry a higher risk of being deprecated or replaced by the model creators. Thus, we recommend using fully-open models (where possible), as well as reporting all relevant parameters used to elicit responses from an LLM, to maximize the chance of reproducibility by other researchers.

A second pitfall is insufficient validation. As emphasized throughout the paper, we cannot recommend the use of LLM-generated norms when the procedure used for obtaining those norms has not been *validated* using a human "gold standard". That is, researchers should ensure that the generation process (including the model, temperature, prompt, and construct of interest) results in judgments that align well with human judgments for words or phrases that do have human judgments. Here, we recommend calculating the correlation between human judgments (e.g., the average of multiple independent human judgments) to LLM-generated judgments; human inter-annotator agreement can be used as a baseline for the degree of alignment expected. We also recommend conducting additional analyses, such as the substitution analysis or error analysis described earlier (Trott, 2024a), which can shed light on issues relating to the *validity* of LLM-generated stimuli; this question is addressed in more detail in Section 7.2 below.

Another crucial pitfall is the possibility of *data contamination*. As discussed earlier, data contamination is when the same data are used to train and evaluate a model. Concretely, data contamination raises challenges for validating an LLM's ability to generate useful psycholinguistic judgments: if an LLM has been trained on concreteness judgments for words like "table" or "freedom", then its ability to *reproduce* those exact judgments may not reflect is ability to produce judgments for words it has not been trained on. There are several approaches to addressing data contamination when validating LLM-generated norms. First, with fully-open models, researchers can conduct an *audit* of the training set to determine whether the data used to evaluate an LLM were included in its training; if they were, a different dataset (i.e., new words with new judgments) should be used. This is (again) an advantage of using fully-open models, and a disadvantage to using commercial models. Second, even with commercial models, researchers can use datasets that were extremely *unlikely* to have been included in a model's training set. The best candidates here would be datasets collected by the researchers themselves. Researchers could also use datasets published after a model's reported "cutoff date", though notably, the cutoff date cannot necessarily be validated or verified. Third, researchers can deploy methods developed to detect whether a particular string was present in the training set (Golchin & Surdeanu, 2023). These methods have the advantage of being usable even for commercial models, but their primary disadvantage is that they are imperfect and may yield false negatives.

## *7.2. Open Questions*

Using LLMs to generate psycholinguistic norms is a relatively novel approach. Therefore, there are a number of open methodological and theoretical questions relating to their use. The primary purpose of this manuscript is to serve as a practical guide; nonetheless, we briefly touch on some of these issues here.

One notable question is *when* and *how* to integrate LLM-generated norms into the psycholinguistic pipeline (e.g., norming data). At the level of individual researchers, this decision will be informed by both empirical analyses (e.g., correlation with a human gold standard) and theoretical concerns (e.g., face validity; see below). Yet as with many methodological questions, we suspect that individual decisions will depend on field-specific consensus about "best practices", such as the use of *p < 0.05* as a threshold for statistical significance. These best practices have yet to be developed. Our recommendation, however, is that LLM-generated norms should currently only be used to *augment* human datasets: that is, researchers should not rely on LLM-generated judgments for stimuli that do not also have human judgments. LLM-generated judgments could be combined with human judgments by averaging them, either with the human mean or by considering them as yet another individual judgment; see Trott (2024b) for additional considerations relating to the efficacy of these approaches.

This connects to a second open question: namely, whether LLM-generated norms exhibit construct validity. As described in Trott (2024a), there are several approaches to addressing this question. One is theoretical, and corresponds roughly to *face validity*: are LLMs the "kind of system" capable of generating responses about a given construct? Here, some constructs (such as part-of-speech) might seem intuitively like more likely candidates than others (such as moral valence). Details of LLM training procedures might be relevant here as well: for instance, LLMs are not currently trained on olfactory or gustatory inputs, so one might decide that they are not suitable for answering questions about taste or smell, except insofar as a researcher's goal is determining whether such information can be gleaned from the distributional statistics of language alone. A second approach is empirical, and relies on validating LLM-generated judgments by comparing them to a human gold standard (see above) or asking whether they correlate with *other* data of interest (such as lexical decision times) in similar ways as the human gold standard. Empirical approaches might also be used to address theoretical questions, such as the degree of *alignment* between the mechanisms underpinning LLM-generated and human-generated responses (Pavlick, 2023).

Another set of open questions relates to the *statistical analysis* of LLM-generated data. At present, psycholinguists deploy a number of statistical techniques developed to address common issues relating to experimental design and data collection (e.g., repeated-measures designs). For instance, non-independence in data (such as multiple responses from the same participant) is typically addressed using mixed effects models that include *random effects* terms intended to account for those sources of non-independence. It is unclear how to conceptualize potential sources of non-independence in LLM-generated data. For example, if a researcher uses the same model to produce 100 responses to the same stimulus with the same prompt (i.e., with temperature > 0), how should those responses be analyzed? Intuitively, the researcher could include random effects terms for the model, prompt, and item, under the assumption that this procedure roughly corresponds to eliciting responses from the same person at different points in time, yet this assumption may not be correct, and a given LLM response may be more analogous to the *average* between multiple human responses (Trott, 2024b). Relatedly, most work on LLM-generated norms has focused on their ability to produce responses that correlate with the human average. It is unclear whether they can be used capture *individual differences* in responses, or whether such an approach is even theoretically sensible.

Finally, the use of LLMs in this capacity raises questions about research ethics and the responsible use of technology in research. We have touched on some of these issues already, such as reproducibility. Yet there are also deep philosophical questions about the use of systems like LLMs in psycholinguistics and the potential "epistemic gaps" this introduces into the scientific pipeline (Messeri & Crockett, 2024). This is why our current recommendation is to keep the "human in the loop", i.e., using LLM-generated norms to *augment* (rather than *replace*) human-generated norms. Further, we note that there is a growing body of literature on the ethical issues relating to the use of AI in scientific practice (Messeri & Crockett, 2024; Agnew et al., 2024; Lin, 2025; Abdurahman et al., 2024; Wang et al., 2025).

## 8. Conclusion and limitations

In this work, we have shown that LLMs may be a useful tool for estimating word characteristics that require human judgments. However, due to the novelty of these models, their rapid adoption by the scientific community, and the (often overlooked) factors influencing the responses they produce, there is a need for practical guides that explain how to fully exploit their potential, avoid common pitfalls, and maximize the chance of reproducibility by other researchers.

To address this need, we have presented a guide that explores both the direct use of base LLMs and the training of specialized models that leverage the prior linguistic knowledge of these base models. We have provided practical guidelines and insights derived from our experience, including prompting strategies, approaches for evaluating results, model configuration and selection, dataset sampling methods, and fine-tuning. These contributions are accompanied by a framework that we believe can serve as a reference for future research in the field (Appendix A). The use of the framework will make the studies replicable and allow the steps taken to be traced, something crucial in this type of research since, as has been seen, some strategies work very well in certain scenarios and not in others. We have also discussed common pitfalls and open questions relating to the use of LLMs in producing psycholinguistic norms. Appendix A also includes a step-by-step guide showing how to execute the framework through the English familiarity with and without fine-tuning using both a commercial model (GPT-4o.mini) and an open-weights model (Llama3.1-8B).

Fine-tuning has emerged as a promising technique for improving model alignment with target tasks, although its effectiveness varies depending on the specific lexical feature and language under consideration. Fine-tuning not only improves model performance in certain cases, but also helps reduce the uncertainty associated with the direct use of an LLM, as it tends to be less sensitive to prompting strategies.

Finally, as described in Section 7, the use of LLM-generated norms has several limitations:

1. **Inscrutability.** LLMs operate as highly complex neural networks with billions of parameters, making their internal reasoning processes effectively opaque. For commercial models, this issue is even more pronounced, as they are provided as black boxes, preventing any inspection of their internal mechanisms, the precise parameters underlying their responses, or the data used to train them.

2. **Uncertainty about optimality**. The results presented are based on months of experimentation with the proposed approach. However, given the lack of interpretability of LLMs, we cannot ensure that our solution is the optimal one. It is likely that future research will introduce alternative strategies that outperform our method.

3. **Dependence on commercial models**. The best performance was achieved with commercial LLMs. This introduces the risk that if these models are discontinued, they will no longer be available for use. Furthermore, there is no guarantee that future versions of LLMs will outperform current ones in psycholinguistic tasks of the kind explored in this paper. Fine-tuning can serve as a strategy to mitigate this risk by retraining models to the specific task.

4. **Variability across languages, characteristics, and models**. We experimented with multiple languages (English, Spanish, German, Dutch), lexical characteristics (familiarity, age of acquisition, valence, arousal, etc.), and models (GPT-4, Gemini 2.5, LLaMA). We found that performance varied considerably across these dimensions. This highlights the importance of having a reliable, human-validated dataset to assess whether model estimations are sufficiently accurate for each specific context.

## Appendix A. Doing the work

Now that we have described the main steps to obtain LLM estimates for psycholinguistic variables, we will elaborate a few examples. In doing so, we will briefly repeat some of the information already given. The reason for this is that we think most readers will either read the first seven sections or this Appendix. They will read the previous sections when they want to get a general idea, and they will want to dive directly into this section when they want to apply the methods to a specific question.

Here, we present a framework and define a methodology based on our experience in the past two years, when we developed techniques for a range of studies involving different word features, languages, and techniques. The text presents practical examples, including code to run the experiments, generated datasets and validation sets (see https://github.com/WordsGPT/psycholinguistics_framework). Importantly, our discussion also includes some dead ends we encountered: configurations that led to useless results.

Figure 1 summarizes the five steps of the methodology. In the first step, validation data is collected which will be used to evaluate the quality of the LLM estimates we obtain. The second step involves selecting the LLM, its configuration, and the prompt design. In the third step, the first estimations are obtained from the chosen LLM. In the fourth step, the validation data is divided into a training and test set. The training data is used to fine-tune the LLM; the test set is used to see how much the fine-tuning improved the estimates provided by the model. Finally, in step 5 the fine-tuned model is used to estimate new words for which no human data is available.

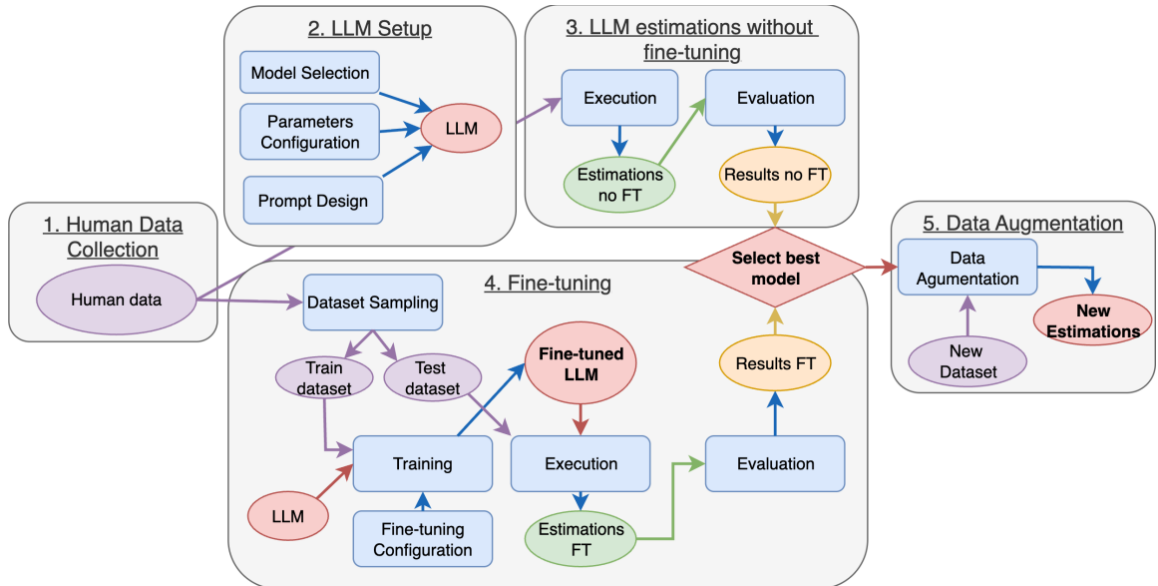

*Figure 1: Estimation of word characteristics with fine-tuned LLMs methodology*

To make this section hands-on, we include a case study on estimating familiarity values in English. We have applied this methodology to other case studies in different languages and with different word characteristics, and we have seen the best results in English.

The advantage of a pipeline protocol is that it reduces the time needed to perform experiments, promotes reproducibility, and reduces the chance of errors, since everything is controlled by configuration files and the execution is automated via scripts. If you want to replicate the results, you must first download the code and install the required Python libraries (step-1), then create a folder for your case study and a "config.yaml" file that will serve as the main configuration file for the experiments (step-2). There is a template of this "config.yaml" file in the code. We have already created the folder "familiarity_english" with all the code and configuration files that can serve as a reference.

A.1. *Collecting validation data*

There is little point in generating LLM estimates of word features, if you do not have data to evaluate them. Computers are extremely good at generating output, but there is little inherent guarantee that the data is of any use for the question you want to address. Even the best documented program with lots of face validity can produce useless output if a small programming error was made in one of the hundreds of lines of code. LLMs can be used to check code and avoid mistakes, but quality control works best if you have a pipeline that always compares the output to what is expected to happen.

The evaluation of predictions is especially important because, although LLMs have proven to be promising tools, their performance varies depending on the type of task. For example, models that seem very powerful in linguistic reasoning can show limitations in seemingly simple tasks like counting letters in words (Conde et al., 2025c). Predicting the performance of an LLM for a specific task is difficult, so having a real and well-annotated criterion dataset becomes an essential requirement to validate its usefulness. If you do not have a validation criterion, you cannot expect people to have faith in the data you produce, and you have no way of optimizing your work.

There are three ways to obtain validation data and we encourage researchers to use all three of them, because this provides them with rich feedback.

### A.1.1. *Human data in the target language*

The first validation criterion consists of human data. If you claim that the estimates you obtain from an LLM have psycholinguistic value, you must show their usefulness to approximate human data. Such data can be human ratings or other performance indices (e.g., reaction and accuracy in a word processing task). If you are lucky, you can find datasets other people already collected. It is worthwhile to do a deep search for these, as each dataset is a goldmine for you. One dataset is good, but two or more are better, because this allows you not only to look at the correlation of your LLM output to the human data, but also to see how much a set of human data correlates with other human data. A low correlation with a human dataset indicates poor LLM performance, but less so if different datasets also show low correlations: in such cases, it becomes clear that the human dataset is suboptimal or that the construct simply exhibits low agreement among humans.

If no human dataset can be found (or if only one can be found), it is necessary to collect new data. Sometimes this dataset can be rather small if the question is very specific. For instance, Martínez, Molero et al. (2025) collected new human data to verify the quality of LLM estimates of valence and arousal for English words. Because all other evidence pointed to a high quality of the estimates, it was enough for the authors to take 100 words with a wide range of valence estimates and 100 words with a large range of arousal estimates and to present them for a rating study to some 20 people each. Correlations between the LLM estimates and the human ratings were higher than 0.9, suggesting that LLM-generated estimates were indeed high quality.

That said, it is best to have human data for a few thousand words. Generally, humans can provide trustworthy ratings of word features at a rate of 1000 per hour (based on Conde et al., 2025b). So, the (time) investment is not too much of an obstacle. The number of raters you need depends on the correlation between the ratings of the different participants. For Likert ratings, these tend to be high (at least 0.2), meaning that most of the time 10-20 raters will be enough to get reliable ratings. A sensible approach is first to collect data from 12 raters and collect the reliability of these ratings. This can be as simple as Cronbach's alpha, available in several R packages (e.g., psych; Revelle, 2015) and in almost any statistical package. The reliability should be larger than .8 and if it is not, it is worthwhile checking whether there may be one or two participants who responded carelessly. You can see this by calculating the correlation between the ratings of each participant with the average rating of the remaining participants (again present in nearly all packages that contain Cronbach's alpha). If a participant responded in line with the other participants, there will be a positive correlation between the participant's ratings and the ratings of the others. If the participant responded carelessly, there will be no correlation between their ratings and those of the others.

If the reliability is smaller than .8 and there are no signs of careless responding, then there is divergence in the responses of the participants and more raters must be tested before the average scale value has the desired reliability[9]. You can estimate how many extra participants you need to test with the Spearman-Brown prophecy formula: n = 0.8*(1-$r_{obtained}$) / $r_{obtained}$*(1-0.8). If your ratings study has a reliability of .6 after 12 participants were tested, you will need 0.8*(1-0.6)/0.6*(1-0.8) = 1.6 as many participants in total or 1.6*12 = 19.2 participants (thus, 8 extra participants to test).

A reliability of 0.8 means that the maximum correlation you can find between your LLM estimate and the human ratings is $\sqrt{0.8}$ = 0.89. Most of the time it will be lower, just like the correlation of your ratings with those of other researchers is unlikely to be higher than 0.8 (the maximum correlation between two variables with less than perfect reliability equals $\sqrt{r_1 * r_2}$; $r_1$ = reliability variable 1, $r_2$ = reliability variable 2). If you find this ceiling level too low, you can test extra participants until you have a reliability of 0.9, but this will require (many) more participants.

When collecting human ratings, it is good to keep in mind that data gathering should not make it easy for participants to use AI. The days when participants could be given a spreadsheet and asked to fill in the ratings are over, unless you can fully trust your participants. Even tools that require participants to click the value for each item online may no longer be fully safe, given that apps exist that automatically generate an LLM answer for multiple-choice questions or Likert scales. One way to check for these is to make sure that the data of participants do not correlate more with a zero-shot LLM from the most popular public tools than with the other participants.

---

[9] This is the case, for instance, with ratings of arousal, where individuals give quite divergent ratings about which words they find arousing.

> For the English familiarity case study, we took existing human data based on the 2,545 words for which there are human familiarity ratings in both the MRC database (Coltheart et al., 1981) and the Glasgow database (Scott et al., 2019). In the step-3) you have to save the dataset in XLSX or CSV format in the "data" folder with at least two columns: the word, and the estimation value.

### A.1.2. Translation of data from another language

A second way to get validation data is by translating rated words from another language. English, Dutch, Spanish and Chinese are languages with large databases of human data (e.g., ratings of age of acquisition, concreteness, valence, arousal, familiarity; or performance-based measures, such as word prevalence and lexical decision responses). Something that works well is to translate words from the source language (e.g., English) into the target language you are interested in (e.g., German) and then back to the source language (English). Words with the same back translation as the original words are words that have very similar meanings in both languages. Forward and backward translation can be automatically (e.g., with Google translate, DeepL, or an LLM).

### A.1.3. Other AI estimates

Finally, there is a chance that some AI-based dataset already exists for the language you are interested in. This is the case, for instance, for word concreteness, valence and arousal, where researchers have calculated estimates for 50+ languages based on semantic vectors (Buechel et al., 2020; Hollis et al., 2017; Plisiecki & Sobieszek, 2023; Solovyev et al., 2022; Thompson & Lupyan, 2018; Want & Xu, 2024).

AI-based estimates do not help you determine the efficacy of your own LLM-generated norms in predicting human judgments, but at least they provide you with information about how good your estimates are relative to the existing AI standards. Ideally, you find that the new estimate outperforms the existing measures, thus correlates more with the validation criteria.

## A.2. LLM Setup

Once you have validation data, you can proceed to generating estimates with the LLM you have for the language you are interested in. Again, there are many choices and decisions to make.

### A.2.1. Model selection

Model selection is a crucial and non-trivial task given the wide variety of options currently available. There are models of different sizes, costs, and task orientations, so choosing the most appropriate one for a specific problem requires careful analysis.

To date, the models that have provided the best results for us are the commercial LLMs GPT-4o and GPT-4o-mini, both from OpenAI. Importantly, by "best" results we mean the estimates that provide the highest correlations with human ratings. So, they are limited to their usage as psycholinguistic norms. Researchers interested in *how* models produce specific estimates and which variables influence them may benefit more from the use of an open-source, research-based model, even if the output correlates less with human ratings. Further, as noted above, we expect that the performance of open-source models will continue to improve in the coming years.

It is important to highlight that the most expensive or powerful commercial models do not always offer the best results. For example, we obtained worse results with GPT-4.5 than with GPT-4o and GPT-4o-mini. The alignment of the model with the cognitive and linguistic characteristics of human language processing is a key factor in achieving good results and is not always linked to the price or the size of the model. In the case of GPT-4.5, the model has been focused on improving programming capabilities, a task distant from natural human language processing. In a similar way, using the new reasoning models does not necessarily provide significant advantages since prediction of word features does not seem to require complex reasoning capabilities. Again, this reinforces the need for validation data in making an evidence-based choice of model.

Another relevant decision is the choice between open weights models and commercial models. Open weights models require somewhat more complex technical setup, since they typically need to be downloaded, installed, and managed on private or public infrastructure. Having said that, increasingly graphical user interfaces, such as Anaconda AI Navigator, are becoming available that reduce the technical skills needed by researchers to deploy open-source models locally on their systems. Alternatively, LLMs can be accessed through providers that facilitate access (for both commercial and open weights models), such as the HuggingFace Endpoints Hub[10]. In contrast, commercial models usually offer access via API, which simplifies use but introduces some limitations. A significant disadvantage of commercial models is the lack of access to weights and intermediate layers, making it difficult to carry out explainability studies. Additionally, there is a risk of losing access if the company decides to change or even deprecate a given model, which interferes with the reproduction and extension of published results. A recent example of this situation is the discontinuation of GPT-4.5 access. Also, a university or other funding source may decide

---

[10] https://huggingface.co/inference-endpoints/dedicated

to stop paying for an outdated version and switch to the new one, whenever it becomes available.

Finally, it is essential to document and keep the exact version of the model used. Model providers often release periodic updates of the same model that alter model behavior. For example, the version "gpt-4o-mini-2024-07-18" indicates the specific variant of GPT-4o-mini deployed on that date.

Model selection is neither simple nor unique. We recommend consulting the literature to identify the most suitable models according to the task's nature and to perform comparative tests with different models and configurations to choose the optimal option. It is also a good idea to keep a time-stamped copy of each validated estimate you obtain, so that you can keep on using it when the model changes or is no longer available.

A golden rule is to never trust estimates that have not been subjected to the validation checks, even if the new estimates are a rerun of a program that worked well in the past. **Validation should be an inherent part of the pipeline you use whenever you obtain AI estimates of psycholinguistic features.**

> Within the English study we selected the commercial model GPT-4o-mini (gpt-4o-mini-2024-07-18, from OpenAI) and the open weights model Llama3.1-8B (from Meta)

### A.3. Collecting LLM estimates

Once you have decided which LLM model you are going to work with, you will need to make a number of decisions about how to collect the estimates. Here we mention the most important ones we've encountered in our research so far.

#### A.3.1. Parameter configuration

The most important parameter for estimating word features within an LLM is temperature. Other parameters can usually be left at their default values unless a specific problem is detected during testing that justifies adjusting them. There are several strategies for configuring temperature:

1) **Temperature = 0.** Setting temperature to 0 aims to make the process as deterministic as possible, thus facilitating the replicability of the experiment. Still, it is good to know that temperature = 0 does not mean absolute determinism because of the stochastic sampling in the commercial models hosted by companies like OpenAI. You can consider this the reliability of the AI estimates. Most of the times, we have found correlations above .96 between two runs. When determining the reliability of the AI estimates, it is important to make sure that the order of items differs between the runs, so that you check to what extent the estimates are influenced by previous estimates made.

2) **Temperature = 0 with logprobs.** This configuration is especially recommended when the task involves predicting an integer that is represented with a single token[11]. In these cases, it is possible to obtain the logarithmic probabilities (logprobs) of the top-k tokens and calculate the weighted average to capture the complete distribution of possible responses, not just the most likely estimate. This approach provides more granular estimates. When the required output is non-integer (e.g., numbers with decimal places), it is not advised to use logprobs because the output is likely to consist of multiple tokens, and obtaining the full log-probability tree is extremely computationally intensive, as its complexity grows exponentially.

3) **Temperature ≠ 0**. In this case, the result you obtain will differ between runs, because a temperature different from 0 indicates that the estimation is not deterministic and was generated using the estimated probabilities of the possible output tokens. In other words, the test-retest reliability of the estimates will be reduced (you can easily test this by calculating Cronbach's alpha of different runs with temperature set at a particular value). When a temperature different from 0 is chosen, you can get more reliable estimates by repeating the experiment a number of times and calculating the average values (just like the average human rating is calculated). Authors may prefer this approach, if they think randomness is an inherent part of estimation and they suspect that the value obtained with temperature = 0 does not correspond to the outcome based on averaging several noisy runs. Again, having validation criteria is a great help here to make informed decisions. A possible strategy is to run the model multiple times with the default temperature and capture statistics such as mean, mode, median, and standard deviation of the predictions obtained as independent samples. This allows capturing the inherent variability of generation with non-zero temperature. If resource limitations prevent multiple runs, it is preferable to use temperature 0 to ensure reproducibility of the experiment.

> The framework is configured with Temperature = 0 and registers the logprobs. It uses the logprobs when possible, i.e., when the output is limited to an integer between 0 and 999.

*A.3.2. Prompt Design*

---

[11] For example, in the GPT-4o tokenizer Integer numbers from 0 to 999 are represented using a single token.

Prompt design consists of providing instructions to the LLM to make the estimation. To some extent, LLMs have made it easy to write prompts, because we can use the same instructions as we give to people. Indeed, a good starting point is to use the instructions used in a classic human rating study (or the rating study you ran to obtain validation data).

At the same time, it is good to keep in mind that small changes in the prompt may affect the quality of the results. For example, model performance may benefit from the inclusion of certain background information in the prompt. Take the following simple prompt:

> 1) *Rate the familiarity with {Word} on a scale from 1 to 7. Return a number.*

This prompt supposes that the model attaches the same meaning to the word familiarity as you do. It also supposes that the model shares your concept of a 7-point scale and how it should be used to extract the maximum of information, given the range that can be expected. For instance, if you only intend to present words likely to be known by people, a good interpretation of answer 1 is "a word I may have come across before but I do not know what it means". Then the entire range of 1 – 7 is used. In contrast, if the stimulus list includes word unlikely to be known by people, the meaning of answer alternative 1 should shift to "a word I've never seen or heard before", in order to include the entire range of values that will be encountered. Also, what is the proficiency level you have in mind: that of an adult native speaker, a child, or a second language learner?

It is a good idea to make the above assumptions explicit by expanding the prompt (as is done in most human rating studies as well). For instance, the following prompt may be better:

> 2) *Complete the following task as a native speaker of English. Familiarity is a measure of how familiar something is. An English word is very FAMILIAR if you see/hear it often and it is easily recognisable. In contrast, an English word is very UNFAMILIAR if you rarely see/hear it and it is relatively unrecognisable. Please indicate how familiar you think this English word is on a scale from 1 (VERY UNFAMILIAR) to 7 (VERY FAMILIAR), with the midpoint representing moderate familiarity. The English word is: "{Word}". Only answer a number from 1 to 7. Please limit your answer to numbers.*

In this example (based on Scott et al., 2019), we are priming the LLM to yield the information we have in mind (remember that the LLM produces the most likely token given the preceding context). We can investigate the effect of the more detailed prompt by obtaining estimates with both prompts. If prompt 2 is better than prompt 1, it will provide estimates that correlate more with the validation dataset(s). If there is no difference between the prompts, we can limit ourselves to the shortest prompt to reduce processing time and overall cost. If prompt 2 is better than prompt 1, we can ask ourselves whether other information may further improve it. For example, would the prompt become better if we add a few examples of words that are expected to get an estimate of 1 and words that are expected to get an estimate of 7? If providing that information does not improve performance (much), it may be better to leave it out, as you are interfering less with the model's training regime. On the other hand, if the extra examples increase the correlations by a nontrivial amount (e.g., +0.05) you may want to include them in the prompt, as it brings the estimates closer to human evaluations.

In addition to *content*, the *format* of the prompt also matters. Some examples of formatting choices include:

1) Asking the model to return both the word and the number (instead of just the number), so that you protect yourself against alignment issues in case the LLM does not provide a value for a few words in the list (which we often see in our experiments).
2) Asking the model to return a JSON with a specific format, including the word: *The output format must be a JSON object. For example: {"Word": "{Word}", "Familiarity": // Familiarity of the word expressed as a number from 1 to 7}* This command helps to manipulate the results later through coded programs (i.e., scripts).
3) Including the word in the prompt without a colon, without quotation marks, and in the middle of the sentence: *[…] Please indicate how familiar you think the English {Word} is on a scale from 1 (VERY UNFAMILIAR) to 7 (VERY FAMILIAR), with the midpoint representing moderate familiarity][…]*
4) Making the prompt in a different language. Unlikely to lead to better results if you want estimates for the English language, but something you may want to try out if you are looking for estimates in a language that did not figure prominently in the training and fine-tuning of the LLM. Sometimes you may get better estimates with English instructions than with instructions in the language you are investigating.
5) Asking for decimal numbers: *Only answer a number from 1 to 7. Please limit your answer to numbers, it may include up to two decimal places.* It is recommended to limit the number of decimal places, as otherwise the model might hallucinate and generate out of range numbers.
6) Asking for an integer from a larger range, e.g. from 10 to 70, so that we can divide the estimate by 10. *[…]Please indicate how familiar you think each English word is on a scale from 10 (VERY UNFAMILIAR) to 70 (VERY FAMILIAR), with the midpoint representing moderate familiarity. […] Only answer a number from 10 to 70. Please limit your answer to numbers.*
7) Adding a few examples to the prompt (few-shot). Including examples with the expected results can help the LLM understand the task better. It is important to provide representative examples that cover all possible values. *[…] For example,*

> "imam" has a familiarity of 2, "theology" has a familiarity of 4, and "fridge" has a familiarity of 7. The English word is: "{Word}" [...].

These are but a few prompts we tried out and the strategies we found useful. The number of possibilities is much larger and possibly unlimited. In addition, a prompt may work well for one task and badly for another, or may perform differently across LLMs.

> In step-4) include all these prompts in TXT files stored in the folder "prompts" with one file per prompt. These files will serve as templates for building all the queries to the LLM. You have to use the *{Word}* expression in the parts where the specific word to estimate would be replaced.

Estimates for a new variable or a new language may benefit from experimentation with multiple options to see what effects they have on the outcome. Of course, this introduces the risk of overfitting (or "prompt" hacking); at the same time, optimal engineering often requires some degree of experimentation (Ward, 1998). Researchers can ameliorate this risk by preserving some amount of "held-out data", which is not used during this experimentation phase—much like a typical "test" set in machine learning. Alternatively, or additionally, researchers could collect entirely new data to validate the estimates.

With well-established languages and variables, a straightforward approach is simply to use the instructions given to humans, i.e., those used in the published article. At the same time, the instructions that are most helpful for humans may not be the instructions that produce the LLM-generated judgments that most strongly correlate with human judgments; thus, some experimentation can be helpful, depending on the researcher's goal.

Future research is likely to limit the search space of effective prompts. So, it is good practice to review the state of the art in search of the best prompts, to be transparent about the strategies you tried out, and to run validation tests when you think you've achieved a major break-through.

### A.3.3. Execution

As we outlined before, it is not a good idea to work with an interactive web interface to obtain LLM estimates of word features, unless the interface used by the researcher allows settings adjustments to variables such as temperature, top P values, and the continuity of past messages.[12] The most user-friendly web interfaces have been developed to make interacting with an LLM intuitive and pleasant, so they do not give you control over the parameters, and the estimates are likely to be affected by what you did before because they shaped the context window within which you are working. This phenomenon, known as context contamination, can introduce unwanted bias in the results, affecting the validity of the generated estimates.

---
[12] Some universities and workplaces have interfaces to GPT with these functionalities.

It is much better to use the API. By using the API, each query is sent as an independent call, where the model always starts "from scratch" without memory of prior inputs. This approach ensures that each prediction is unconditioned by previous estimates, which is key to guaranteeing consistency and reproducibility of the experiment. The use of the API in commercial models is not free; it is charged per input and output token. You must register and obtain an API key. There is a wide range of prices among models, and access to LLMs varies internationally, as do costs. Nonetheless, for most western-based university researchers, today GPT-4o-mini costs $0.15 per million input tokens and $0.60 per million output tokens, while GPT-4o costs $2.50 (input) and $10 (output) per million tokens. One million tokens is equivalent to about 750,000 words in English, including both the input prompt and the model's response. Any other language different from English, will be more expensive as the models represent words with more tokens[13]. There are online calculators that estimate the cost of running experiments[14]. For instance, running prompt (2) on 100,000 words would cost $2.19 with GPT-4o-mini and $35 with GPT-4o in English. The price is not always aligned with quality and depends on the provider. From our experience, GPT-4o-mini is sufficient for linguistics tasks, certainly if fine-tuning is used.

LLM providers offer different interaction modes through their API. The most direct method is synchronous processing, where an individual request is sent and the response is received immediately. This option is useful for small tasks, quick validation, or cases where interactive inspection or debugging of results is needed. However, when working with large volumes of data, it is more efficient to use the asynchronous mode, known as batch processing. In this mode, the user submits a file with all the questions to be processed, and the provider returns the responses within a maximum of hours (typically 24). This approach does not require keeping local infrastructure running and it is usually cheaper.

Despite its advantages, batch processing requires some consideration. Some entries may fail during the process, so it is essential to verify that all requested estimates have been generated. Additionally, it is important to check that all responses have the correct format, since occasionally the model may deviate from the instructions, producing unexpected outputs or structural errors. For example, it has been observed that when asked in the prompt to also return the word, the model changes the input word to another word in the output. In such cases, it is recommended to rerun the failed entries, either by making new individual calls or generating a new batch exclusively with those examples.

---

[13] https://help.openai.com/en/articles/4936856-what-are-tokens-and-how-to-count-them
[14] https://gptforwork.com/tools/openai-chatgpt-api-pricing-calculator

> The framework is prepared to run in batch mode using the API. In step-5) configure the "apis.env" with your API key (you have a template of this file in "apis_example.env"). If you are going to share the code from your experiments, make sure not to upload your API_key, as someone else could use it in your name. Then in step-6) you have to configure the config.yaml, including the name of the experiment, the XLSX with your dataset, the name of the column where the words to estimate are, the file with the prompt to evaluate, and the LLM to use. As an example, for the first prompt the content would be:
> *experiments:*
>   *familiarity_english_v01_short_prompt:*
>     *dataset_path: "Glasgow_MRC_joint_norms_inner_join_english.xlsx"*
>     *dataset_column: "Word"*
>     *prompt_path: "english_v01_short_prompt.txt"*
>     *model_name: "gpt-4o-mini-2024-07-18"*
>
> In step-7) run the program "python3 prepare_experiment.py <EXPERIMENT_PATH> <EXPERIMENT_NAME>" replacing EXPERIMENT_PATH with the name of the case study folder ("familiarity_english"), and EXPERIMENT_NAME with the name of the experiment ("familiarity_english_v01_short_prompt"). As a result, a file with all the queries to be made to the model will be generated in a folder called "batches". Check in this file that the generated prompt is what you expect. If your dataset contains more than 50,000 words, several files will be generated due to OpenAI API limitations.
> Then (step-8) run "python3 execute_experiment.py <EXPERIMENT_PATH> <EXPERIMENT_NAME>" to send your petitions to the model's API. In batch mode, it may take up to 24 hours to complete the results, meanwhile you can turn off your computer as nothing runs on it. To download the results, you have to access the OpenAI website – batches section.
> In step-9) download the results and save them in a folder called "results." The file with the results contains a lot of information related to the model execution. To make the results manageable, run "python3 generateResults.py," which will process all the results and save them in a new XLSX file for future analysis.

*A.3.4.* Evaluation

Evaluation consists of assessing how accurate the estimates are compared to the validation measures. We recommend calculating both Spearman and Pearson correlations (Conde et al., 2025a). Most of the time they will not differ much, in which case you know the correlations can be relied on. Sometimes, however, the correlations will differ considerably. This is most likely when the distribution of a variable is asymmetric, for example when you have a variable with many low values and a few high values (e.g., how much is each word related to smell?). To understand differences between Pearson and

Spearman variables, it is good to know that the Pearson correlation gives more weight to extreme values (i.e., those at the low and the high end), whereas the Spearman correlation gives more weight to (possibly small) differences around the mode. A higher Pearson correlation than Spearman correlation indicates that the correlation largely depends on words at the high and/or low end of the scale. A higher Spearman correlation than Pearson correlations indicates that the correlation is mainly situated in the middle of the scale with deviating items at the extremes.

To make things more tangible, Figure 2 presents some results for the prompts we discussed with respect to English familiarity. The data are based on the 2545 words for which there are human familiarity ratings both in the MRC database (Coltheart et al., 1981) and Glasgow database (Scott et al., 2019). The Pearson correlation between these two datasets is $r = .79$; the Spearman correlation is $\rho = .80$ These are the baselines to compare the LLM estimates with. GPT estimates show similar results in its best configurations, with Spearman correlations of $\rho = .79$ (MRC) and $\rho = .82$ (Glasgow), and Pearson correlations of $r = .76$ (MRC) and $r = .81$ (Glasgow). As mentioned before, the way we ask the model is important. For example, the short prompt without explanation (prompt-v01) produced the worst estimations. In fact, for many words, it provided additional text before giving the estimate. In the analysis, we did not observe a difference between a prompt asking the model to return both the number and the word versus only the number. Marking the word in the prompt is relevant, since without a colon and quotation marks, the Spearman correlation dropped by .12. Rescaling strategies had a negative impact here, and writing the prompt in Spanish did not show a negative effect. What induced a significant improvement, up to .10 in correlation, was using logprobs. The best prompt when considering both datasets was the few-shot one. We included three reference words: "imam" (familiarity of 2), "theology" (4), and "fridge" (7). It is important to note that these results are not the ground truth. In other scenarios, we have obtained very different outcomes, and what works here may not work in other cases. With Llama3.1-8B (the open-weights model) we obtained worse results. In fact, there are large differences between the different strategies tested. The one that worked best was the standard prompt with logprobs, but it was .20 points below its counterpart in GPT-4o-mini. It is also noteworthy that the use of logprobs was also very effective in Llama, since the correlation without logprobs in Llama3.1-8B was .30 points lower.

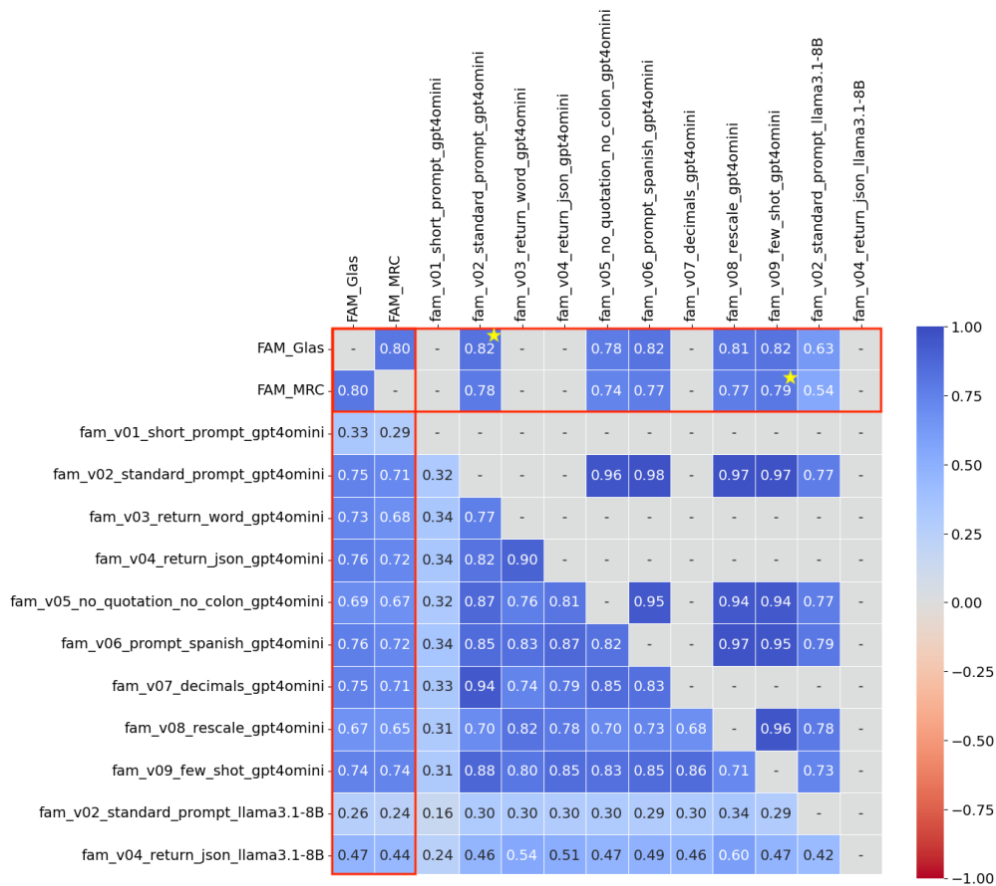

Figure 2: `Spearman` correlations of the GPT and Llama prompts with Glasgow and MRC English familiarity databases (over the 2545 common words). Above the diagonal: correlations calculated with logprobs (when possible); below the diagonal: correlation using the output of the model.

### A.4. Fine-tuning the model

Depending on the quality of the LLM estimates, you may look into improving them by fine-tuning the model. In fine-tuning, the model receives feedback about the expected values, so that the weights can be adjusted and the output more closely resembles the desired one. Such fine-tuning is not needed if the model already produces estimates on par with what you observe in humans (as in Figure 2), but is worth considering when the estimates from step 3 are below expectation. If is not a fail-safe solution, but we found that the effect sometimes can be quite pleasing.

#### A.4.1. Dataset sampling

Fine-tuning requires data to feed to the model. Most of the time, the data will be a part of the validation data you collected in step 1. An alternative strategy is to collect new human data based on the results of step 3. For instance, we could extract a sample of interesting words based on the familiarity estimates obtained in Figure 2. These can be words that had different ratings in MRC and Glasgow, words for which the different prompts gave divergent estimates, words with a big difference between the familiarity estimate and word frequency. These are interesting words to collect human data for, in order to bring the model more in line with human opinion.

At the same time, it is good to keep in mind that it is not a good idea to feed the model only with exceptional cases, as this risks distorting the estimates for the rest of the words. A good strategy is to combine some interesting new words with words for which the model already gives good output.

The number of words that are needed for fine-tuning depends on the noise in the data. For reaction times (which are inherently noisy) you will need more feedback than for ratings on a Likert scale. We found that not much gain is made after feedback on 2000 words for Likert ratings of AoA (Sendín et al., 2025), whereas 5000 words were needed for fine-tuning a model to predict reaction times in a lexical decision task (Martínez et al., 2025).

The more data you have to fine-tune the model, the more closely the model's output will resemble the validation criterion. LLMs already have a lot of linguistic knowledge based on their massive pretraining and all that is needed is to bring the output better aligned with that knowledge. So, even if you only have a small number of data to feed back to the model, it can still be worthwhile to test what effect it has.

Importantly, to evaluate the effect of fine-tuning you need a dataset that was not used in the tuning itself. There is little point looking at whether fine-tuning improved the estimates for the words used as feedback, as this will always be spectacular (the model learns what it should return for these particular words). The big question is how much better fine-tuning makes the estimates for *new* words, a process called cross-validation. Cross-validation requires that you collect data for new words after the fine-tuning, or that you split your available dataset into at least two subsets: one for training and another for evaluation. In classical machine learning problems, a standard split of 80% for training and 20% for evaluation is adopted. However, in the context of fine-tuning, this ratio can be more flexible and it may be better to split 60-40, so that you have enough data to test the outcome of the fine-tuning. If you only have data for 1000 words, it is not really reassuring if your quality test is based on 200 words only. In general, we recommend aiming for data for at least 3000 words, 1800 of which can be used for training and 1200 for testing. With sample sizes of more than 1000, the results are unlikely to differ much between different random splits of the data into a training and testing set. In case of doubt, it is always possible to fine-tune a model more than once and look at the extent to which the results depend on the specific training and test samples extracted.

The quality, diversity, and coverage of the dataset are important variables to consider, in addition to the size. Regarding sampling strategies, the most straightforward approach is to simply take a random selection of all stimuli. This works well if each stimulus is equally interesting, but will reduce the information extracted if only a subset of words is of interest. For example, you could take a random sample of a list of 120,000 English lemmas for familiarity estimates. However, given that more than half of these words are unlikely to be known to your participants, you are throwing away much information, because all these words end up in the lowest bin. It may be more interesting, to limit the words to words that are likely to be known to some people, for example based on their frequency or on previous word prevalence information.

Stimulus selection for finetuning also has a dark side, because you are excluding potentially interesting information. For instance, Sendín et al. (2025) finetuned a Spanish model to 3000 AoA ratings obtained by Alonso et al. (2015). These ratings were limited in two ways. First, the words had been selected such that most words were expected to be acquired in the first 12 years of life. Second, participants were asked to categorize all words learned after the age of 11 years to a single category of 11+. This meant that the fine-tuned model could only provide information for early acquired words. All late acquired words got an estimate of 10-11 years old, in line with the fine-tuning regime the model was given.

It is therefore important to always be aware of possible biases and limitations in the training set and to test the fine-tuned estimates on a validation sample of stimuli. With regard to cross-validation, it is important to keep in mind that the value of cross-validation decreases with each additional test you perform. A mistake researchers often make is to try out dozens of cross-validation analyses and keep the best one. Cross-validation involves applying your *chosen* analysis path to a *new* dataset to see how well it generalizes. That is why it is a good idea to always keep one data sample until the very end of your pipeline (or collect new data at that point).

Fine-tuning can also have unexpected positive outcomes. For instance, in experiments estimating word familiarity in German, we observed that fine-tuning did not change much for words typically used in psychological studies, but we observed that the fine-tuned model was much more sensitive to spelling errors in the words (something that often happens in corpus analysis). Whereas the model before finetuning tended to give rather high estimates to recognizable words with small spelling errors, it gave low estimates after having received feedback on 2000 correctly spelled German words (this included the use of capitals for nouns, something typical for German).

### *A.4.2.* Fine-tuning prompts

Because feedback is given to the model after each estimate, the prompt used for fine-tuning can be shorter than the one used for an untuned model. There is no need anymore to include the scale end or to give examples of extreme words. Even the definition of the variable

("familiarity" in our example) can be dropped, because the model gradually learns what you want on the basis of the feedback.

Other decisions remain, such as whether you ask for both the word and the estimate, whether you use logprobs or increase the range of the scale, and so on. For instance, it is not clear what the consequences are if you give feedback about ratings with decimal places. One of the consequences we saw is that this seems to reduce the granularity of the estimates relative to the untuned estimates based on logprobs. The model seemed to restrict the number of outputs it gave. More work is needed here.

Finally, the fine-tuning parameters (e.g., batch size, number of epochs, learning rate) must be selected. Platforms like OpenAI provide default settings that are adapted automatically to the training dataset. In our research, we usually use these default settings, but it is crucial to report the parameters that were ultimately chosen to ensure the experiment's reproducibility.

### A.4.3. Training and evaluation

A model is fine-tuned by first giving the prompt, asking for an estimate. Afterwards, the model is informed about the output that was expected. This is done by providing the correct output as a target, allowing the model to adjust its internal parameters through backpropagation (the standard method for training neural networks), minimizing the error between its prediction and the desired result.

Once the model has seen all the training stimuli, it must be saved, and the fine-tuned model can then be used to provide estimates for the test stimuli. Evaluation of the fine-tuned model is done by looking at how much more the estimates of the fine-tuned model correlate with the validation criteria for the test stimuli. You can also look at the effect for the trained stimuli, but this effect will always be rather large and says nothing about the quality of estimates for new stimuli.

In case of doubt, it is always a good idea to collect some new human validation data. Two advantages of this strategy are that the new findings cannot be contaminated by information already available on the internet (and possibly included in the training of the model) and that you can limit the stimuli to those that are most informative to test the difference between the untuned and the fine-tuned model (e.g., words for which the estimates differ a lot).

The final evaluation must be carried out using a dataset that has not been used in previous stages of training. This evaluation dataset must be kept completely independent to ensure an objective and unbiased assessment.

Following the English familiarity case study, we decided to split the 2545-item dataset into a training set for fine-tuning (1500 entries) and a test set for validation (1045 entries). The sampling strategy was a random split. We tested three different estimations for fine-tuning: based on the Glasgow norms, the MRCnorms , and the mean of the Glasgow and MRC norms. This resulted in three different models: ft_v01_glas, ft_v01_mrc, and ft_v01_glas_mrc_mean, respectively. The fine-tuning instruction used was v02_standard_prompt.

> To reproduce the fine-tunings with the framework, you have to complete the following steps:
>
> First, (step-10) add your fine-tuning to the "config.yml". Specify the input file with the inputs and reference estimates, the percentage of inputs you want to use for training normalized to one (if the input dataset contains only the training data, set this to 1), the name of column where the reference value is located, the prompt to use for fine-tuning, the base model to be fine-tuned, and the name of the new model. For example, for fine-tuning with the Glasgow estimates, use the following configuration:
>
> *finetuning:*
>   *familiarity_english_ft_v01_glas:*
>     *ft_dataset_path: "1_train_random_MRC_Glas.xlsx"*
>     *train_split_percentage: 1*
>     *random_state: 42*
>     *answer_column: "FAM_Glas"*
>     *prompt_path: "english_v02_standard_prompt.txt"*
>     *dataset_finetune_name:*
> *"Glasgow_MRC_joint_norms_inner_join_english_ft_v01_glas.jsonl"*
>     *model_name: "gpt-4o-mini-2024-07-18"*
>     *especial_suffix: "familiarity_english_ft_v01"*
>
> After that (step-11) execute the program "python3 create_finetuning_dataset.py <EXPERIMENT_PATH> <EXPERIMENT_NAME>" to build the file with all the estimates, and in step-12) run "python3 execute_finetuning_dataset.py <EXPERIMENT_PATH> <EXPERIMENT_NAME>" to train the new model. Again, before executing the training it is good practice to check that the file with the fine-tuning data is correct. The framework is programmed with the default OpenAI fine-tuning settings. As before, the fine-tuning does not run on your computer, so you can turn it off. Fine-tuning may take several hours, and once finished, you can access the OpenAI website – fine-tuning section to obtain the model and the final training parameters. The new model is saved on OpenAI's servers and can be used by referencing it.
>
> Finally, step-13) consists of running the evaluation dataset with the fine-tuned model, following the same steps described in step-6 to step-9.

Figure 3 shows the results obtained when fine-tuning the GPT and Llama Model. In the case of GPT, the best results for each database occur when the model is fine-tuned using its own estimates, reaching a Spearman correlation improvement of 0.10 (Glasgow) and 0.09 (MRC), as we can expect. Additionally, fine-tuning with the other database had a small improvement of 0.03 (Glasgow) and 0.04 (MRC). The best balance was achieved when using the mean of the Glasgow and MRC norms, which yielded improvements of 0.09 (Glasgow) and 0.08 (MRC), making it the best-performing model. This model achieves a higher correlation than the one observed between Glasgow and MRC. The fine-tuned model can be understood as an intermediate point between both datasets. In the case of the Llama model, we obtained a substantial improvement on Pearson correlation of 0.22 on Glasgow and 0.30 on MRC, achieving performance that is similar to, though slightly worse than, the GPT models. This indicates that fine-tuning helps improve estimations in models that perform poorly in their base version.

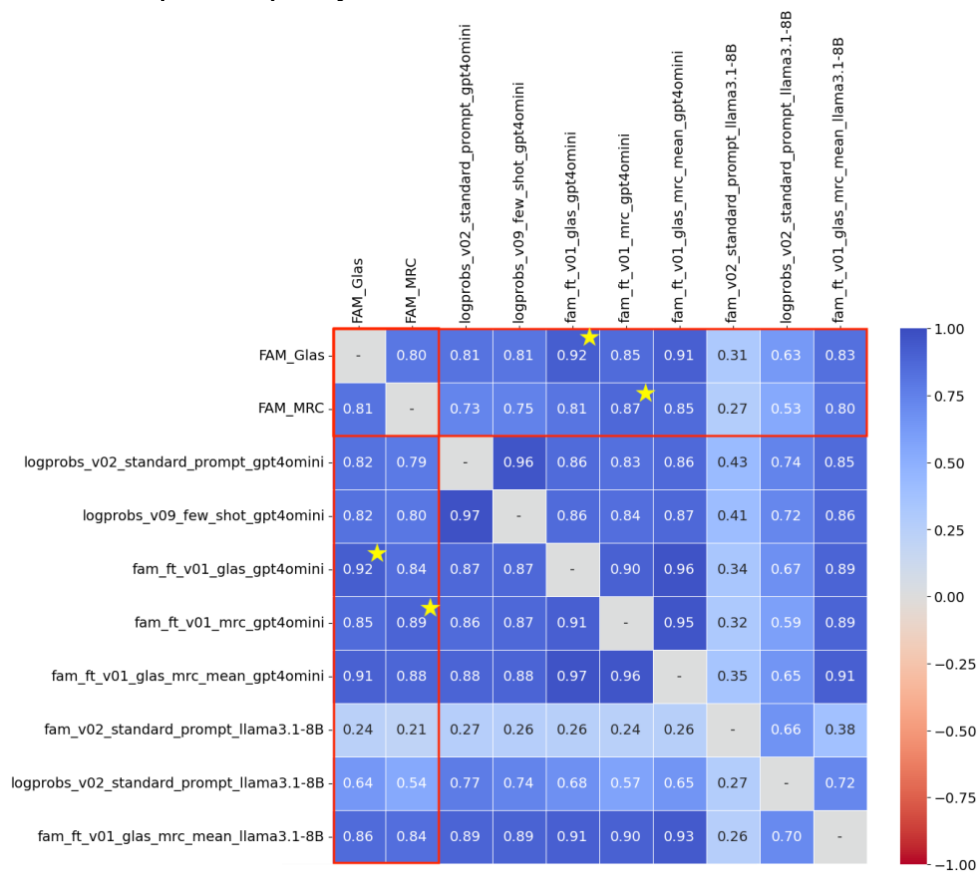

Figure 3: Correlations of the GPT and Llama fine-tuned models with Glasgow and MRC English familiarity databases, v02_logprobs and v09_logprobs prompt versions without fine-tuning, over the 1045 validation words. Above the diagonal: Pearson correlations; below the diagonal: Spearman correlation.

### A.5. Data Augmentation

Once you are happy with the outcome of the (fine-tuned) LLM you are using, you can extend the estimates to all words of interest. In English, this can be as many as 150,000 words; in languages with many inflected or derived forms and compound nouns written as single words, this number can be larger than 2 million (and potentially reaching a few 100 million possible word forms), as any corpus analysis in German, Czech or Korean will attest.

The procedures for augmenting a dataset are the same as those used in the initial generation of estimates and in fine-tuning. It is good practice to maintain rigorous control over all steps carried out and to document everything to ensure reproducibility. This includes recording the versions of the model used, the prompts employed, the parameter configurations, as well as any pre- or post-processing applied to the generated data.
It is also important to remain aware (and to inform users) about limitations in the augmented data. The main limitation will be that you cannot be sure about the quality of the estimates for word structures not present in the fine-tuning dataset. Above we saw the example of age of acquisition, where you cannot be sure about the estimates for words acquired after a certain age. Another example could be when your model was fine-tuned with lemmas only. In that case you have no guarantee that the estimates are also valid for inflected or derived forms. The same is true for compound words. If they were not present in the fine-tuning you cannot be sure of the quality of the estimates for these words.

The advice is always the same: In case of doubt, collect some new human validation data. This data can be gathered in a very focused way (thus need not be too onerous to collect) and in addition can later be used to further improve the fine-tuning of the model. Indeed, the real contribution of AI to psycholinguistics is not that we replace research on human processing with research on AI processing, but that we leverage LLMs to improve our research on human language processing.

It is also a good idea to keep a time-stamped, well-documented version of each validated word list. If this list contains estimates from a fine-tuned model, it is a good idea to also include the estimates before the fine-tuning. In that way, any potential user can verify what the effect of the fine-tuning was.

> To create new data just run again steps 6-9 with the best model and configuration.

## Declarations

- Funding: This research was supported by Programa Propio UPM, the FUN4DATE (PID2022-136684OB-C22) and SMARTY (PCI2024-153434) projects funded by the Spanish Agencia Estatal de Investigación (AEI) 10.13039/501100011033, by the Chips Act Joint Undertaking project SMARTY (Grant no. 101140087) and by the Cotec Foundation. Access to the GPT models was provided by the OpenAI researcher access program.
- Conflicts of interest: The authors ran the studies independently and do not expect any financial gain from them.
- Ethics approval: The studies did not involve people and followed the General Ethical Protocol of the Faculty of Psychology and Educational Sciences at Ghent University. Therefore, they need no explicit approval from the Faculty.
- Consent to participate: The studies did not collected new data from people.
- Consent for publication: All authors consent.
- Availability of data and materials: All data and materials are available at https://doi.org/10.5281/zenodo.17141353.
- Code availability: The Python framework and code used for running the experiments is available at https://github.com/WordsGPT/psycholinguistics_framework.
- Authors' contributions: All authors contributed to the development of the ideas examined in this experiment. The framework was developed by the team in Madrid and Milano. The Ghent group took responsibility for the statistical analysis. Groups from Hong Kong and United States contributed to the state of the art, recommendations, lessons learned, and the writing.

## References

Abdurahman, S., Atari, M., Karimi-Malekabadi, F., Xue, M. J., Trager, J., Park, P. S., ... & Dehghani, M. (2024). Perils and opportunities in using large language models in psychological research. PNAS nexus, 3(7), pgae245.

Achiam, J., Adler, S., Agarwal, S., Ahmad, L., Akkaya, I., Aleman, F. L., ... & McGrew, B. (2023). Gpt-4 technical report. arXiv preprint arXiv:2303.08774.

Agnew, W., Bergman, A. S., Chien, J., Díaz, M., El-Sayed, S., Pittman, J., ... & McKee, K. R. (2024, May). The illusion of artificial inclusion. In Proceedings of the 2024 CHI Conference on Human Factors in Computing Systems (pp. 1-12).

Alonso, M. Á., Díez, E., & Fernandez, A. (2016). Subjective age-of-acquisition norms for 4,640 verbs in Spanish. Behavior Research Methods, 48(4), 1337-1342.

Alonso, M. A., Fernandez, A., & Díez, E. (2015). Subjective age-of-acquisition norms for 7,039 Spanish words. Behavior research methods, 47(1), 268-274.

Arnett, C., Chang, T. A., & Bergen, B. K. (2024). A bit of a problem: Measurement disparities in dataset sizes across languages. *arXiv preprint arXiv:2403.00686*.

Arnett, C., Rivière, P. D., Chang, T. A., & Trott, S. (2024). Different tokenization schemes lead to comparable performance in spanish number agreement. *arXiv preprint arXiv:2403.13754*.

Arnett, C., & Bergen, B. K. (2024). Why do language models perform worse for morphologically complex languages?. *arXiv preprint arXiv:2411.14198*.

Biderman, S., Schoelkopf, H., Anthony, Q., Bradley, H., O'Brien, K., Hallahan, E., ... & Van Der Wal, O. (2023, July). Pythia: a suite for analyzing large language models across training and scaling. In Proceedings of the 40th International Conference on Machine Learning (pp. 2397-2430).

Binz, M., Akata, E., Bethge, M., Brändle, F., Callaway, F., Coda-Forno, J., ... & Schulz, E. (2025). A foundation model to predict and capture human cognition. Nature, 1-8.

Bommasani, R., Klyman, K., Kapoor, S., Longpre, S., Xiong, B., Maslej, N., & Liang, P. (2024). The Foundation Model Transparency Index v1. 1: May 2024. arXiv e-prints, arXiv-2407.

Botarleanu, R. M., Watanabe, M., Dascalu, M., Crossley, S. A., & McNamara, D. S. (2024). Multilingual Age of Exposure 2.0. International Journal of Artificial Intelligence in Education, 34(4), 1353-1377.

Bricken, et al. (2023). Towards Monosemanticity: Decomposing Language Models With Dictionary Learning. Transformer Circuits Thread.

Brysbaert, M., Martínez, G., & Reviriego, P. (2025). Moving beyond word frequency based on tally counting: AI-generated familiarity estimates of words and phrases are an interesting additional index of language knowledge. Behavior Research Methods, 57:28.

Buechel, S., Rücker, S., & Hahn, U. (2020). Learning and evaluating emotion lexicons for 91 languages. arXiv preprint arXiv:2005.05672.

Cevoli, B., Watkins, C., & Rastle, K. (2022). Prediction as a basis for skilled reading: Insights from modern language models. Royal Society Open Science, 9(6), 211837.

Comanici, G., Bieber, E., Schaekermann, M., Pasupat, I., Sachdeva, N., Dhillon, I., ... & Shan, Z. (2025). Gemini 2.5: Pushing the frontier with advanced reasoning, multimodality, long context, and next generation agentic capabilities. arXiv preprint arXiv:2507.06261.

Conde, J., González, M., Grandury, M., Reviriego, P., Martínez, G., & Brysbaert, M. (2025a). Psycholinguistic Word Features: a New Approach for the Evaluation of LLMs Alignment with Humans. In Proceedings of the Fourth Workshop on Generation, Evaluation and Metrics (GEM²), 8–17, Vienna, Austria. Association for Computational Linguistics.

Conde, J., Martínez, G., Grandury, M., Arriaga-Prieto, C., Haro, J., Schroeder, S., Hintz, F., Reviriego, P. & Brysbaert, M. (2025b). Updating the German psycholinguistic word toolbox with AI-generated estimates of familiarity, concreteness, valence, arousal, and age of acquisition. Preprint

Conde, J., Martínez, G., Reviriego, P., Gao, Z., Liu, S., & Lombardi, F. (2025c). Can ChatGPT Learn to Count Letters?. Computer, 58(3), 96-99.

Davies, S. K., Izura, C., Socas, R., & Dominguez, A. (2016). Age of acquisition and imageability norms for base and morphologically complex words in English and in Spanish. Behavior research methods, 48(1), 349-365.

Golchin, S., & Surdeanu, M. (2023). Time travel in llms: Tracing data contamination in large language models. arXiv preprint arXiv:2308.08493.

Grattafiori, A., Dubey, A., Jauhri, A., Pandey, A., Kadian, A., Al-Dahle, A., ... & Vasic, P. (2024). The llama 3 herd of models. arXiv preprint arXiv:2407.21783.

Green, C., Kong, A., Brysbaert, M., & Keogh, K. (2025). Crowdsourced and AI-generated Age of Acquisition (AoA) Norms for Vocabulary in Print: Extending the Kuperman et al.(2012) norms.

Hinojosa, J. A., Rincón-Pérez, I., Romero-Ferreiro, M. V., Martínez-García, N., Villalba-García, C., Montoro, P. R., & Pozo, M. A. (2016). The Madrid Affective Database for Spanish (MADS): Ratings of dominance, familiarity, subjective age of acquisition and sensory experience. PloS one, 11(5), e0155866.

Hollis, G., Westbury, C., & Lefsrud, L. (2017). Extrapolating human judgments from skip-gram vector representations of word meaning. Quarterly Journal of Experimental Psychology, 70(8), 1603-1619.

Hu, J., & Levy, R. (2023, December). Prompting is not a substitute for probability measurements in large language models. In Proceedings of the 2023 Conference on Empirical Methods in Natural Language Processing (pp. 5040-5060).

Huang, X., Zhu, W., Hu, H., He, C., Li, L., Huang, S., & Yuan, F. (2025). Benchmax: A comprehensive multilingual evaluation suite for large language models. arXiv preprint arXiv:2502.07346.

Hussain, Z., Binz, M., Mata, R., & Wulff, D. U. (2024). A tutorial on open-source large language models for behavioral science. Behavior Research Methods, 56(8), 8214-8237.

Jain, S., & Wallace, B. C. (2019, June). Attention is not Explanation. In Proceedings of the 2019 Conference of the North American Chapter of the Association for Computational Linguistics: Human Language Technologies, Volume 1 (Long and Short Papers) (pp. 3543-3556).

Jones, C. R., & Bergen, B. (2024). Does word knowledge account for the effect of world knowledge on pronoun interpretation?. *Language and Cognition*, *16*(4), 1182-1213.

Kuperman, V., Stadthagen-Gonzalez, H., & Brysbaert, M. (2012). Age-of-acquisition ratings for 30,000 English words. Behavior research methods, 44(4), 978-990.

Liesenfeld, A., & Dingemanse, M. (2024, June). Rethinking open source generative AI: open-washing and the EU AI Act. In Proceedings of the 2024 ACM Conference on Fairness, Accountability, and Transparency (pp. 1774-1787).

Lin, Z. (2025). Six fallacies in substituting large language models for human participants. Advances in Methods and Practices in Psychological Science, 8(3), 25152459251357566.

Long, L., Wang, R., Xiao, R., Zhao, J., Ding, X., Chen, G., & Wang, H. (2024, August). On LLMs-Driven Synthetic Data Generation, Curation, and Evaluation: A Survey. In Findings of the Association for Computational Linguistics ACL 2024 (pp. 11065-11082).

Martínez, G., Conde, J., Reviriego P. & Brysbaert, M. (2025). Generating lexical decision times with large language models: Dynamic use of megastudy data. Preprint

Martínez, G., Molero, J. D., González, S., Conde, J., Brysbaert, M., & Reviriego, P. (2024). Using large language models to estimate features of multi-word expressions: Concreteness, valence, arousal. Behavior Research Methods, 57(1), 5.

Messeri, L., & Crockett, M. J. (2024). Artificial intelligence and illusions of understanding in scientific research. Nature, 627(8002), 49-58.

Moreno-Martínez, F. J., Montoro, P. R., & Rodríguez-Rojo, I. C. (2014). Spanish norms for age of acquisition, concept familiarity, lexical frequency, manipulability, typicality, and other variables for 820 words from 14 living/nonliving concepts. Behavior research methods, 46(4), 1088-1097.

Niu, Q., Liu, J., Bi, Z., Feng, P., Peng, B., Chen, K., ... & Liu, M. (2024). Large language models and cognitive science: A comprehensive review of similarities, differences, and challenges. arXiv preprint arXiv:2409.02387.

Pavlick, E. (2023). Symbols and grounding in large language models. Philosophical Transactions of the Royal Society A, 381(2251), 20220041.

Peeperkorn, M., Kouwenhoven, T., Brown, D., & Jordanous, A. (2024). Is temperature the creativity parameter of large language models?. arXiv preprint arXiv:2405.00492.

Petrov, A., La Malfa, E., Torr, P., & Bibi, A. (2023). Language model tokenizers introduce unfairness between languages. Advances in neural information processing systems, 36, 36963-36990.

Plaza, I., Melero, N., del Pozo, C., Conde, J., Reviriego, P., Mayor-Rocher, M., & Grandury, M. (2024). Spanish and llm benchmarks: is mmlu lost in translation?. arXiv preprint arXiv:2406.17789.

Plisiecki, H., Sobieszek, A. (2024) Extrapolation of affective norms using transformer-based neural networks and its application to experimental stimuli selection. Behavior Research Methods 56, 4716–4731. https://doi.org/10.3758/s13428-023-02212-3

Revelle, W., & Revelle, M. W. (2015). Package 'psych'. The comprehensive R archive network, 337(338), 161-165.

Scott, G. G., Keitel, A., Becirspahic, M., Yao, B., & Sereno, S. C. (2019). The Glasgow Norms: Ratings of 5,500 words on nine scales. Behavior Research Methods, 51(3), 1258-1270.

Sendín, E., Conde, J., Reviriego, P., Haro, J., Ferré, P., Hinojosa, J. A., & Brysbaert, M. (2025). Combining the power of large language models with finetuning based on strategically collected human ratings: A case study about age-of-acquisition estimates of Spanish words. Preprint. 10.13140/RG.2.2.27255.12967

Solovyev, V., Islamov, M., & Bayrasheva, V. (2022, November). Dictionary with the evaluation of positivity/negativity degree of the Russian words. In International Conference on Speech and Computer (pp. 651-664). Cham: Springer International Publishing.

Song, S., Hu, J., & Mahowald, K. (2025). Language models fail to introspect about their knowledge of language. arXiv preprint arXiv:2503.07513.

Staub, A. (2025). Predictability in language comprehension: Prospects and problems for surprisal. Annual Review of Linguistics, 11, 17–34.

Templeton, et al. (2024), Scaling Monosemanticity: Extracting Interpretable Features from Claude 3 Sonnet, Transformer Circuits Thread.

Thompson, B., & Lupyan, G. (2018). Automatic estimation of lexical concreteness in 77 languages. In Proceedings of the Annual Meeting of the Cognitive Science Society (Vol. 40). Available at https://escholarship.org/content/qt7dz7k3k1/qt7dz7k3k1.pdf

Trott, S., & Bergen, B. (2023). Word meaning is both categorical and continuous. *Psychological Review*, *130*(5), 1239.

Trott, S. (2024a). Can large language models help augment English psycholinguistic datasets?. *Behavior Research Methods*, *56*(6), 6082-6100.

Trott, S. (2024b). Large language models and the wisdom of small crowds. *Open Mind*, *8*, 723-738.

Turpin, M., Michael, J., Perez, E., & Bowman, S. (2023). Language models don't always say what they think: Unfaithful explanations in chain-of-thought prompting. Advances in Neural Information Processing Systems, 36, 74952-74965.

Wang, A., Morgenstern, J., & Dickerson, J. P. (2025). Large language models that replace human participants can harmfully misportray and flatten identity groups. Nature Machine Intelligence, 1-12.

Ward, N. (1998). Artificial intelligence and other approaches to speech understanding: Reflection on methodology. Journal of Experimental and Theoretical Artificial Intelligence, 10, 487-493.

Wang, T., & Xu, X. (2024). The good, the bad, and the ambivalent: Extrapolating affective values for 38,000+ Chinese words via a computational model. Behavior Research Methods, 56(6), 5386-5405.

Wendler, C., Veselovsky, V., Monea, G., & West, R. (2024, August). Do llamas work in english? on the latent language of multilingual transformers. In Proceedings of the 62nd Annual Meeting of the Association for Computational Linguistics (Volume 1: Long Papers) (pp. 15366-15394).

Wilcox, E. G., Pimentel, T., Meister, C., Cotterell, R., & Levy, R. P. (2023). Testing the predictions of surprisal theory in 11 languages. Transactions of the Association for Computational Linguistics, 11, 1451-1470.

Yarkoni, T., & Westfall, J. (2017). Choosing prediction over explanation in psychology: Lessons from machine learning. Perspectives on Psychological Science, 12(6), 1100-1122.47